\documentclass{article}
\usepackage{ijcai24}
\usepackage{times}  
\usepackage{helvet}  
\usepackage{courier}  
\usepackage[hyphens]{url}  
\usepackage{graphicx} 
\usepackage{caption} 
\usepackage{booktabs}
\usepackage{amsmath}
\usepackage{amsthm}
\usepackage{tabularx}
\usepackage{algorithm}
\usepackage{algorithmic}
\usepackage{color}
\usepackage{soul}
\usepackage[hidelinks]{hyperref}
\usepackage[utf8]{inputenc}
\usepackage[switch]{lineno}
\usepackage{newfloat}
\usepackage{listings}

\DeclareCaptionStyle{ruled}{labelfont=normalfont,labelsep=colon,strut=off} 
\floatstyle{ruled}
\newfloat{listing}{tb}{lst}{}
\floatname{listing}{Listing}
\title{Gyroscope-Assisted Motion Deblurring Network}
\author{
    SiMin Luan, \textsuperscript{\rm 1}
    Cong Yang, \textsuperscript{\rm 2}
    Zeyd Boukhers, \textsuperscript{\rm 3}
    Xue Qin, \textsuperscript{\rm 1}  
    Dongfeng Cheng,
    Wei Sui,
    Zhijun Li, \textsuperscript{\rm 1}
    \affiliations {
        \textsuperscript{\rm 1} Harbin Institute of Technology\\
        \textsuperscript{\rm 2} Soochow University\\
        \textsuperscript{\rm 3} Fraunhofer Institute for Applied Information Technology FIT \\
        luansiminiot@gmail.com, cong.yang@suda.edu.cn, zeyd.boukhers@fit.fraunhofer.de, qinxue@me.com, dongfengncsu@gmail.com, wei.sui@horizon.cc, lizhijun\_os@hit.edu.cn
        }
}

\usepackage{bibentry}

\begin{document}

\maketitle

\begin{abstract}
    \newcommand{\red}[1]{\textcolor{black}{#1}}
Image research has shown substantial attention in deblurring networks in recent years. Yet, their practical usage in real-world deblurring, especially motion blur, remains limited due to the lack of pixel-aligned training triplets (background, blurred image, and blur heat map) and restricted information inherent in blurred images. This paper presents a simple yet efficient framework to synthetic and restore motion blur images using Inertial Measurement Unit (IMU) data. Notably, the framework includes a strategy for training triplet generation, and a Gyroscope-Aided Motion Deblurring (GAMD) network for blurred image restoration. The rationale is that through harnessing IMU data, we can determine the transformation of the camera pose during the image exposure phase, facilitating the deduction of the motion trajectory (aka. blur trajectory) for each point inside the three-dimensional space. Thus, the synthetic triplets using our strategy are inherently close to natural motion blur, strictly pixel-aligned, and mass-producible. Through comprehensive experiments, we demonstrate the advantages of the proposed framework: only two-pixel errors between our synthetic and real-world blur trajectories, a marked improvement (around 33.17\%) of the state-of-the-art deblurring method MIMO on Peak Signal-to-Noise Ratio (PSNR).
\end{abstract}

\newcommand{\red}[1]{\textcolor{black}{#1}}
\section{Introduction}
\label{s:intr}
Motion blur, which actively occurs in wheeled and legged robots, autonomous vehicles, and hand-held photography, is one of the most dominant sources of image quality degradation in digital imaging, impacting the robustness of industrial applications~\cite{koo2022survey}. Over the past decades, the research on image deblurring has made considerable strides~\cite{nayar2004motion,chen2008robust,cho2009fast}. In particular, integrating neural networks has remarkably enhanced the effectiveness of deblurring (from blurry to sharp images) techniques~\cite{zhang2020deblurring,kupyn2019deblurgan,zhang2022deep}. However, the performance of these studies hinges heavily on the availability of high-quality blurred datasets and their corresponding ground truth images. For instance, pixel-aligned training triplets (background, blurred image, and blur heat map) are required to train a model for various related tasks.

\begin{figure}[t!]
\centering
\includegraphics[width=1\columnwidth]{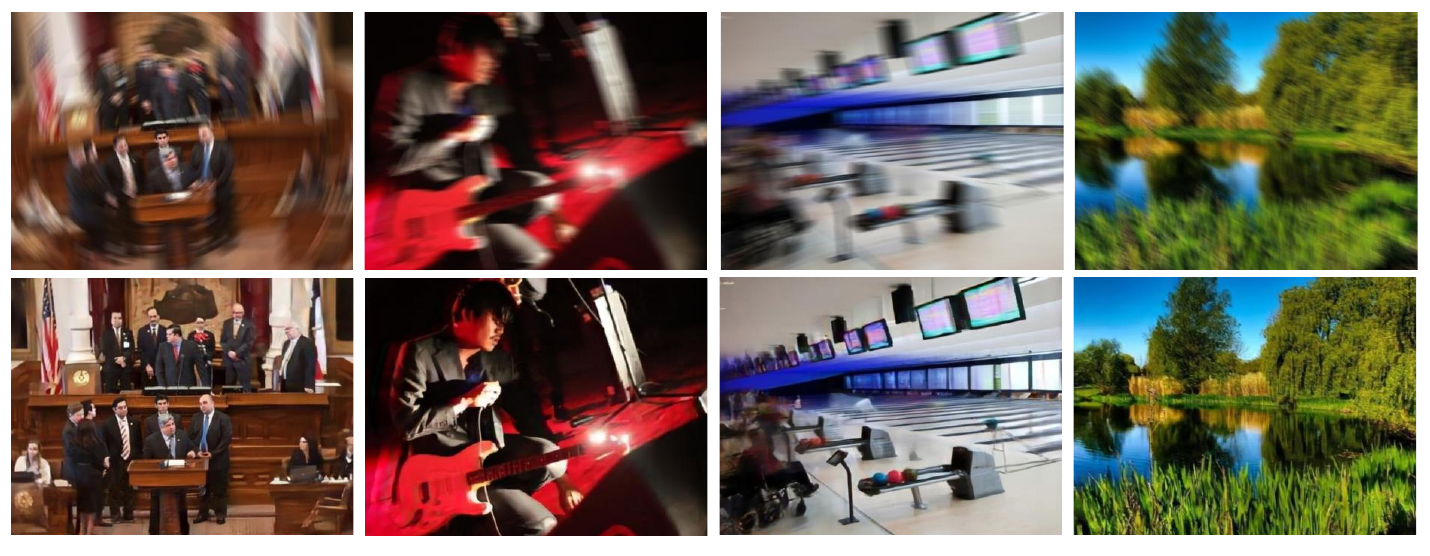}
\caption{Sample results from our proposed framework: The input blurred images (\textbf{top}), and after deblurring (\textbf{bottom}).}
\label{fig6}
\vspace{-1em}
\end{figure}
\begin{figure}[t!]
\centering
\includegraphics[width=1\columnwidth]{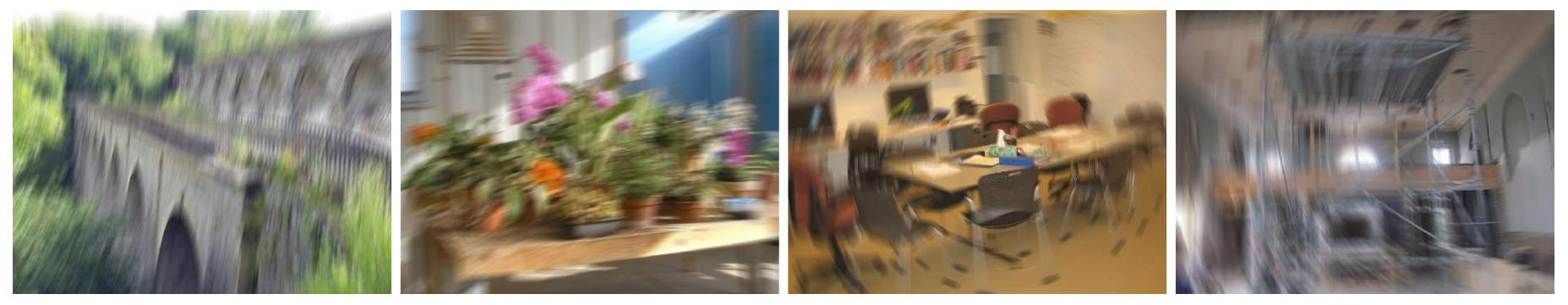}
\caption{Synthetic blur images using our strategy. The blurs are variant and close to the real-world, e.g., rotation blur.}
\label{fig9}
\vspace{-1em}
\end{figure}
Current methods for collecting blurred datasets involve synthesizing blurred images from clear ones. However, existing synthesis approaches assume that the blur trajectory of each pixel (or region within a grid) is the same, which starkly contrasts with reality. In practice, camera-shaking speed and pose are varied (e.g., camera rolling), and motion blur trajectories on each pixel are coherently different. To overcome this, Mustaniemi \textit{et al.} proposed using IMU to synthesize blurred images~\cite{mustaniemi2019gyroscope}. However, they assume the blurred trajectory is a straight line far from the actual situation. As a result, the primary challenge lies in accurately calculating the blurred trajectory for each image pixel. The motion blur presented in Fig.~\ref{fig6} (top) cannot be accurately synthetic with existing methods. In short, such a phenomenon brings two critical challenges to motion blur synthesis and removal: (1) How to effectively simulate real-world blur effects, and (2) how to leverage blur trajectory information to aid image deblurring.

\begin{figure}[t]
\includegraphics[width=1\columnwidth]{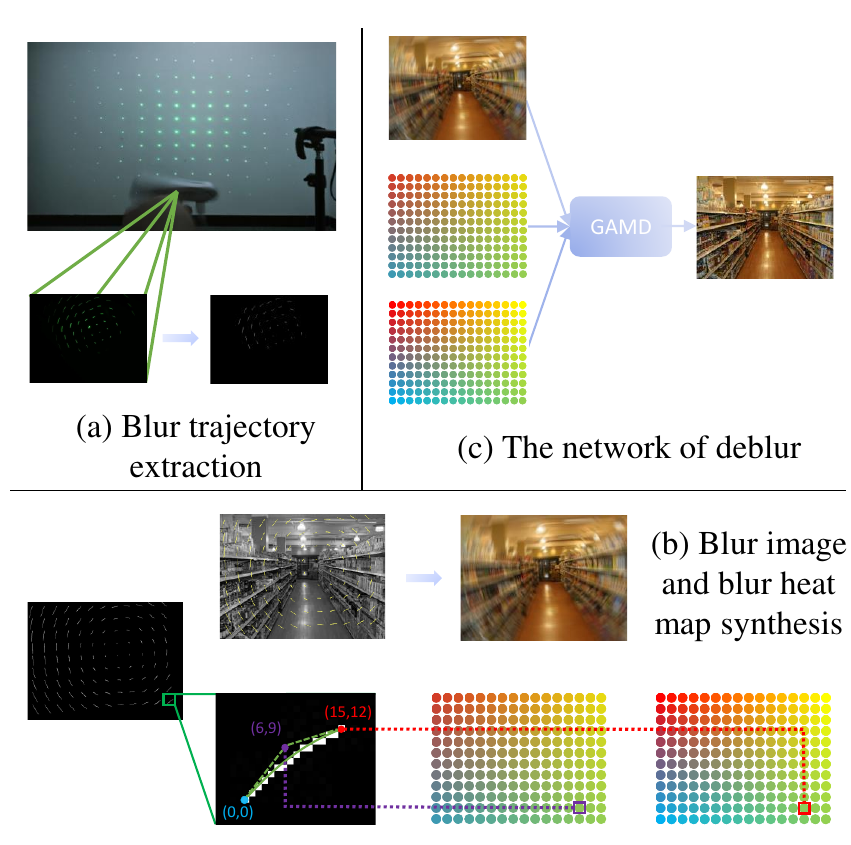}
\caption{(a) Obtain blurred image trajectories and corresponding IMU data. (b) Use our strategy to synthesize training triplets. (c) GAMD deblurring network. }
\vspace{-1.5em}
\label{fig7}
\end{figure}

Given the challenges above, we introduce a framework to leverage blur trajectory on blur image synthetic and deblurring. The framework includes a strategy for training triplet (background, blurred image, and blur heat map) generation and a Gyroscope-Aided Motion Deblurring (GAMD) network for blurred image restoration. We synthesize and restore motion blur images using Inertial Measurement Unit (IMU) data. The rationale is that through harnessing IMU data, we can determine the transformation of the camera pose during the image exposure phase, facilitating the deduction of the motion trajectory (aka. blur trajectory) for each point inside the three-dimensional space. As a result, for the first challenge, our proposed synthetic strategy can fully use the trajectory map generated by the IMU. Specifically, we record the accurate image blur trajectory and the corresponding IMU data by shaking the camera to shoot the laser matrix (see Fig.~\ref{fig7} (a)). The matrix is then employed to guide each pixel to synthesize blurred images according to different trajectories, so the generated blurred images are inherently close to real-world motion blur (see Fig.~\ref{fig7} (b)). Fig.~\ref{fig9} demonstrates that our proposed strategy can generate realistic blurred images in which each pixel is finely shifted according to the trajectory of the motion blur. In particular, our proposed approach can also efficiently create various rotating blurs, which are often omitted from traditional methods.

For the second challenge, we propose a GAMD network based on FPN (Feature Pyramid Networks) ~\cite{ghiasi2019fpn} (see Fig.~\ref{fig7} (b) and (c)) makes full utilize of the blur heat map, that is derived from the blur trajectory, to guide the deblurring in fine-grained level. As shown in Fig.~\ref{fig6},  blurred details are properly restored using our proposed GAMD network. Experiments in Section~\ref{s:exper} show that our framework improves the peak signal-to-noise ratio (PSNR) by about 33.17\% over the state-of-the-art MIMO~\cite{cho2021rethinking}.

In summary, we fully utilise blur trajectory for motion blur image synthesis and restoration. Thus, our contributions are twofold: (1) We introduce an efficient strategy for flexibly modelling blurred images to build pixel-aligned and close-to-real-world training triplets. We also generate a new publicly available dataset (namely IMU-Blur, 8350 triplets, far more than existing datasets in quality and quantity) with our method. (2) We introduce a novel network, GAMD, by learning image features and blur heatmaps to guide image deblurring. Comprehensive experiments demonstrate the accuracy and effectiveness of our proposed framework.
\section{Related Works}
\label{s:relate}

We provide a succinct review of existing works on generating blur datasets and deblurring methodologies. For a more thorough treatment of these topics, recent compilations by Koh~\cite{koh2021single} and Zhang~\cite{zhang2022deep} offer sufficiently good reviews.

\subsection{Motion Blur Collection}
\label{s:related:dataset}
Existing methods are broadly divided into direct simulation and physical acquisition. For the first one, the seminal work by Sun \textit{et al.} synthesizes motion blur by convolving the entire image with a fixed kernel~\cite{sun2013edge}. However, it does not consider the variation of the blur kernel for each pixel. Rim \textit{et al.} propose using gyroscope data to generate realistic blur fields~\cite{mustaniemi2019gyroscope}. Although spatial transform convolution is performed, it ignores the relationship between IMU and blur trajectories. Thus, it cannot generate blurry images that closely resemble natural environments.

For the second one, the images are usually acquired using high-speed cameras. For instance,  Levin \textit{et al.} collects blurred photos by capturing the projected images on a wall while shaking the camera~\cite{levin2009understanding}. Though it better approximates real-world blur, the background and the blurry images are not pixel-aligned. Moreover, its efficiency and quantity are limited for network training. The GoPro dataset~\cite{nah2017deep} is currently the most extensively utilized in deblurring research, created by capturing high-speed video and synthesizing blurred images (via clear frame average). Some real-world blur datasets (with blurred and sharp images)~\cite{rim2020real} were captured using dual cameras simultaneously. Nevertheless, these datasets comprise images from a single scene, potentially limiting the generalization of network training.

Unlike existing approaches, we use the camera IMU data to synthesize blur trajectories and extract large-scale and pixel-aligned blur images from backgrounds. Thus, our proposed strategy is more efficient, and the generated training triplets have higher quality in pixel alignment, close to real-world blur and diversity.

\begin{figure}[!t]
\centering
\includegraphics[width=1\columnwidth]{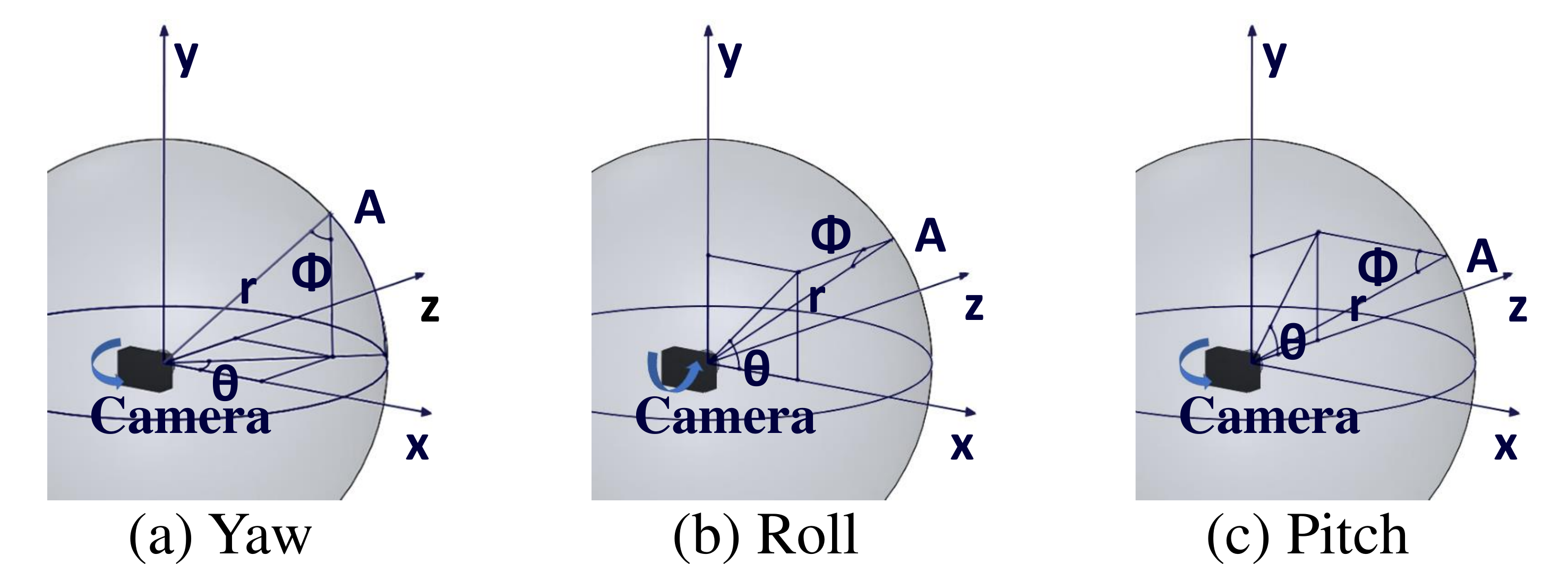} 
\vspace{-1.5em}
\caption{Coordinate system transformation corresponding to different camera motions of Yaw, Roll, and Pitch.}
\label{fig1}
\vspace{-0.5em}
\end{figure}

\begin{figure}[!h]
\centering
\includegraphics[width=1\columnwidth]{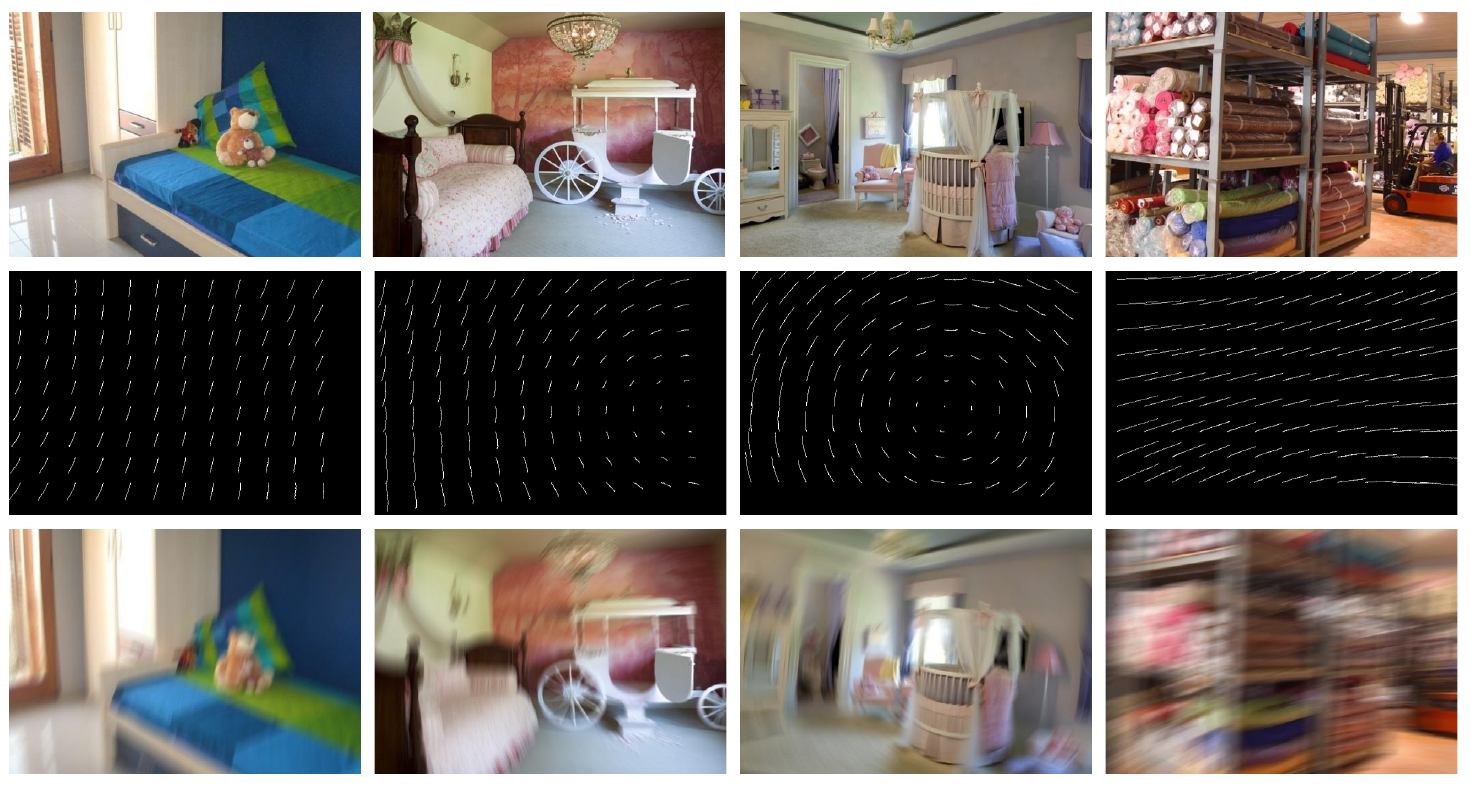} 
\vspace{-1.5em}
\caption{The clear shape image (top), blurred trajectory calculated from the IMU data (middle), and blurred image (bottom) synthesized by the method in this paper. }
\label{fig2}
\vspace{-1.5em}
\end{figure}

\subsection{Deblurring}
\label{s:related:deblur}
Image deblurring is a long-standing problem in computer vision. Earlier deblurring methods focused primarily on uniform deblurring~\cite{cho2009fast,fergus2006removing}. However, these methods often fall short as real-world images typically exhibit non-uniform blurring, with varying blur kernels across different image regions. In the deep learning era, numerous CNN-based blind deblurring approaches have been proposed. For instance, the generative adversarial network DeblurGAN~\cite{kupyn2018deblurgan} uses blurred and sharp image pairs to train a conditional GAN for deblurring. SRN~\cite{tao2018scale} employs recurrent networks to extract features between images at multiple scales.
Similarly, MIMO-UNet~\cite{cho2021rethinking} extracts image features at various scales and merges these features to complete the deblurring task. However, these blind deblurring algorithms often fail to achieve satisfactory results due to the lack of blur-related prior on region, intensity, and motion trajectory. To overcome this problem, we feed a heatmap (converted from IMU data) into the network to guide deblurring. \red{At the same time, we found a slight error between the blur prior data generated by the IMU and the ground truth. These errors are proportional to the size of the image pixels and are negligible on coarse-grained images. Therefore, we use the image pyramid network structure to learn blur trajectories and blur image features. Increasing the proportion of coarse-grained features can enhance the image deblurring effect.} Experiments show that our method can effectively improve deblurring accuracy, achieving a state-of-the-art performance.

\section{Method}
\label{s:meth}
We first review the principles of blurred image generation in the real world and, based on this, propose a method for triplet generation. Next, we detail how GAMD utilizes blur trajectories to guide image deblurring. Note that the blurred images we discuss here are all produced by camera movement.

\subsection{Blur Trajectories}
\label{s:meth:blurtra}
Motion blur is mainly due to the relative motion between the camera and the scene within a limited exposure time. Suppose there is a three-dimensional space point $A$ in the camera lens. When the camera moves, the blur trajectory of point $A$ on the image is superimposed by its imaging position within the exposure time. In this stage, there is a mapping relationship between the point $A$ on the image and the pose change of the camera. At the same time, the camera pose at each moment is obtained based on the IMU data. So, we can derive the blur trajectory of point $A$ through the IMU and use the blur trajectory to synthesize the corresponding blurred image. Table~\ref{tab3} summarizes all the variables and their descriptions.

\begin{table}[!t]
\centering
\caption{Crucial symbols and their descriptions.}
\begin{tabularx}{.95\columnwidth}{l|X}
    \hline
    Symbol & Description  \\
    \hline
    $U, V$ & The coordinates of the mapped position of point $A$ on the camera\\
    $X, Y, Z$ & The coordinates of point $A$ under the camera coordinate axis\\
    $\boldsymbol{I}_t(x)$ & The instantaneous clear image captured by the camera at time $t$\\
    $f_x,f_y$ & The focal length of the camera in the $x(u)$ and $y(v)$ directions\\
    $\boldsymbol{a}_t$ & An array of $U, V$ at time $t$\\ 
    $\boldsymbol{B}(x)$ & Blurred image captured by the camera\\ 
    $\boldsymbol{B'}(x)$ & Blurred image synthesized by our method\\
    $\boldsymbol{P}_t$ & The pose of the camera at time $t$\\
    $t$ & When $t=0$, the above symbols represent the initial state\\
    
    \hline
\end{tabularx}
\label{tab3}
\vspace{-0.5em}
\end{table}

Next, we separately discuss the effects of camera translation and camera rotation on the transformation of the imaging position of point $A$ on the image. When the camera only moves in translation, there is the following equation:

\begin{equation}
\begin{aligned}
    & \boldsymbol{P}_{t2} = \boldsymbol{P}_{t1} + \boldsymbol{T} \\
    & Z_{t1}\boldsymbol{a}_{t1} = \boldsymbol{KP}_{t1} \\
    & Z_{t2}\boldsymbol{a}_{t2} = \boldsymbol{KP}_{t2} \\
\end{aligned}
\quad .
\label{meth:translation}
\end{equation}
where $\boldsymbol{T}$ is the camera motion's translation matrix, $\boldsymbol{K}$ is the camera's intrinsic matrix, and $Z_t$ is the depth of point $A$ from the camera at time $t$. Since the camera's exposure time is generally in the range of tens of milliseconds, the translational displacement of the camera is minimal during this period. We can assume that $Z_{t1} = Z_{t2}$, and calculate the pixel displacement of point $A$, ($\boldsymbol{a}_{t2}-\boldsymbol{a}_{t1}$) as $\boldsymbol{KT}/Z$. This equation shows that as the point $A$ is located at a deeper position ($Z$ becomes larger, $\boldsymbol{K}$ and $\boldsymbol{T}$ remain constant), the effect of camera translation on pixel motion becomes smaller. A real-world analogy is when driving a car and looking at an obstacle on the side of the road - the closer this obstacle is to the vehicle, the faster it moves.

Therefore, when the camera captures distant objects, we can ignore the effect of camera translation on image motion blur. Regarding camera rotation, when shooting, all light passes through the centre of the camera lens. Inspired by this phenomenon, we use spherical polar coordinates to represent points in space. This alternative representation facilitates problem-solving. The rotation of the camera consists of Pitch, Roll and Yaw. Since the camera exposure time is very short, we can approximate the blur trajectory during this period as the vector sum of the blur trajectories generated by each movement. The following discussion explores motion blur caused by three types of motion.

In yaw-induced motion blur, the spatial coordinate representation is depicted in Fig.~\ref{fig1}. The $XYZ$ coordinates can be expressed as follows:
\begin{equation}
\begin{aligned}
    X &= r\sin{\phi}\cos{\theta} \\
    Y &= r\cos{\phi} \\
    Z &= r\sin{\phi}\sin{\theta}
\end{aligned}
\quad .
\label{meth:polarcoordinate}
\end{equation}
where $r, \theta, \phi$ are marked in Fig.~\ref{fig1} (a). Combining Eq.~\ref{meth:polarcoordinate} and camera imaging principles, the pixel coordinates of the spatial point mapped onto the image are:
\begin{equation}
\begin{aligned}
    U &= \frac{f_xX}{Z} = \frac{f_x}{\tan{\theta}} \\
    V &= \frac{f_yY}{Z} = \frac{f_y}{\sin{\theta}\tan{\phi}}
\end{aligned}
\quad .
\label{eq:projection}
\end{equation}

It's evident from the above Eq.~\ref{eq:projection} that the variable $r$ cancels out. When capturing images, the distance $r$ between the target object and the centre of the camera lens does not influence the projected pixel position. From the spherical coordinate system, $\phi$ remains constant as the camera undergoes yaw. Only $\theta$ changes, and this change can be determined through the camera's gyroscope readings. For a given pixel point $A$ (known $U$, $V$), we can derive the corresponding $\phi$ and $\theta$ values under various blur conditions using  Eq.~\ref{eq:projection} and use the gyroscope to acquire $\Delta\theta$.

\begin{equation}
\begin{aligned}
    {\Delta}U &= f_x\left(\frac{1}{\tan{\theta}} - \frac{1}{\tan{(\theta + \Delta\theta)}}\right) \\
    {\Delta}V &= \frac{f_y}{\tan{\phi}}\left(\frac{1}{\sin{\theta}} - \frac{1}{\sin{(\theta + \Delta\theta)}}\right)
\end{aligned}
\quad .
\label{eq:projectiondelta}
\end{equation}

\begin{figure*}[!t]
\centering
\includegraphics[width=0.9\textwidth]{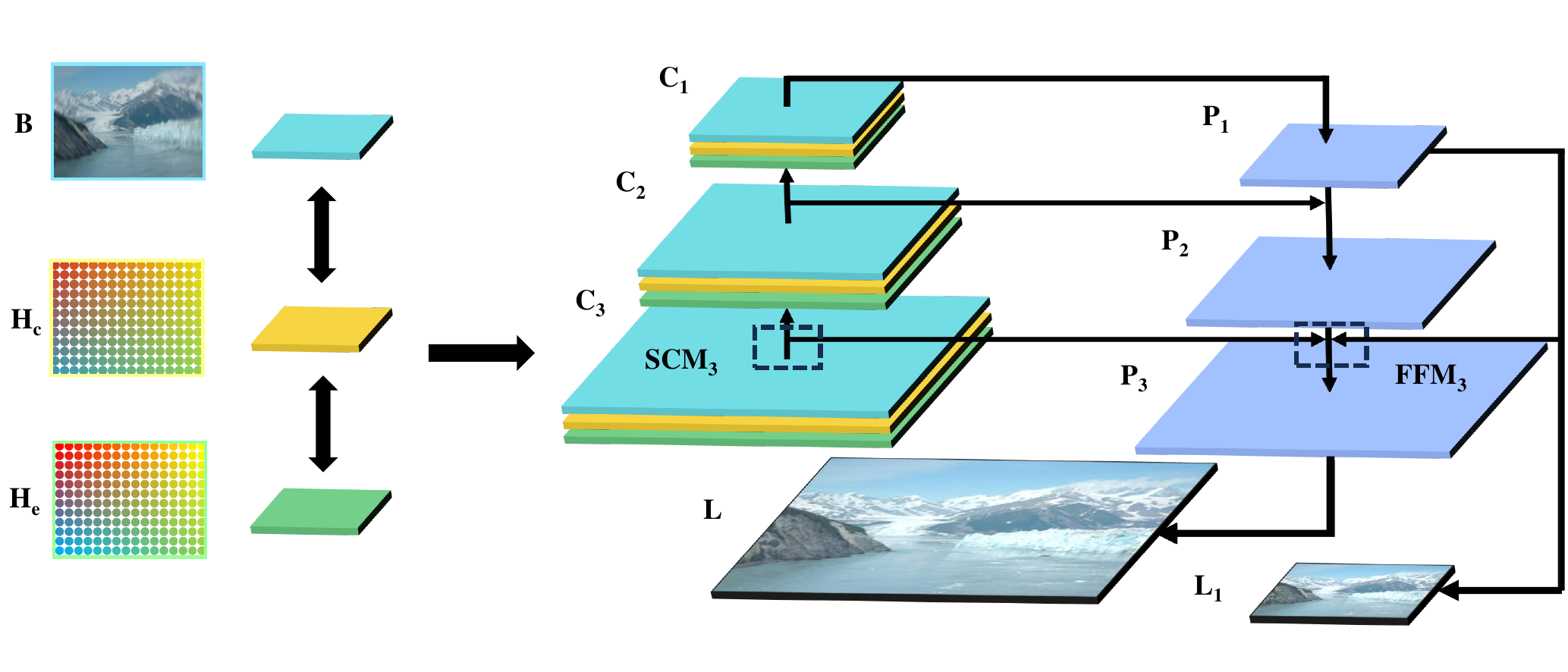} 
\vspace{-1em}
\caption{ The architecture of GAMD network. $B$ is the blurred image, $H_c$ and $H_e$ are the data heat maps of the control points and endpoints of the blur trajectory of all pixels, respectively, and $L$ is the clear image after restoration.}
\vspace{-1.5em}
\label{fig3}
\end{figure*}

Due to space limitations, the remaining two ambiguous conditions are discussed in detail in the appendix. Using Eq.~\ref{eq:projectiondelta}, we can calculate the difference in a blur for each pixel and stack these calculations to calculate the projected point of the target point on the new image. Connecting these projected points is the approximate blur trajectory of point $A$. This process needs to iteratively update the latest projection point coordinates and gyroscope information from the moment the camera exposure starts until the end of the direction. Our experiments show that the gyroscope data of our device corresponds to six sets of data in each image frame. The pixel origin for each pixel can be computed by computing the corresponding blur trajectory node along seven possible trajectories. These seven nodes connect to form a curve that outlines the trajectory. However, the accuracy of blur trajectories is inherently limited by the frequency of gyroscope data acquisition. Increased accuracy requires higher frequency gyroscope data collection.

After calculating the pixel trajectories of the four corners of the image, the projective transformation can be performed on the entire clear image, and the images collected by the camera at each moment of the exposure time can be calculated. A blurred image close to the natural environment can be obtained by superimposing these images. The imaging principle of the camera involves collecting photons during the exposure time. This process can be integrated using the following formula:
\begin{equation}
\begin{aligned}
   \boldsymbol{B}(x) = \lambda\int_{0}^{\tau}\boldsymbol{I}_{t}(x)dt
\end{aligned}
\quad .
\end{equation}

\noindent where $\boldsymbol{B}(x)$ represents the blurred image captured by the camera, $\lambda$ is the normalization factor, $\boldsymbol{I}_{t}(x)$ is the clear image captured at timestamp $t$, and $\tau$ denotes the camera's exposure time. Given that our gyroscope can collect six sets of data during the camera's exposure time, the camera moves at different speeds during each interval, resulting in uneven blur in the image. The composite blurred image $\boldsymbol{B'}(x)$ is derived as follows:

\begin{equation}
\begin{aligned}
    \boldsymbol{B'}(x) &= \frac{1}{m}\sum_{j=1}^{m}\frac{1}{n_j}\sum_{i=1}^{n_j}\boldsymbol{I}_{ij}(x) \\
    n_j &= \max(l_{jk})
\end{aligned}
\quad .
\end{equation}
where $m$ is the number of gyroscope-acquired measurements during the camera's exposure time, and the blur intensity varies across different stages. $I_{ij}(x)$ denotes the image after the $i$-th perspective transformation of the original image in the $j$-th stage. It defaults to the first frame's clear image at the start of the exposure. For clarity, we select four points on the map as original points in the first frame and calculate the blur trajectories of these points using the gyroscope. We compute the corresponding perspective transformation matrix by selecting corresponding points on the four blur trajectories and derive the new instantaneous clear image $I_{ij}(x)$. $l_{jk}$ represents the pixel length of the blur trajectory corresponding to the $k$-th pixel in the $j$-th stage. To optimize the blur effect, we set the length of the longest blur trajectory as $n_j$. We synthesize a blurred vision by superimposing these images, as depicted in Fig.~\ref{fig2}.

\subsection{Deblurring Network}

\subsubsection{Blur Trajectory in the Network}
Unlike traditional neural networks that mainly focus on blurry images, we refer to heatmaps generated based on fine-grained blur trajectory data. To do so, we use the camera IMU data to calculate the blur trajectory for each pixel to create the heatmap. In practice, it is challenging to convert blur trajectories into heatmaps since the mathematical image of most trajectories is intractable. For this, we introduce a Bezier curve, a parametric curve defined by discrete ``endpoints" and ``control points" that describe the blur trajectory for each pixel. Here, we briefly summarize the advantages of Bezier curves:
\begin{itemize}
\item Flexibility: Bezier curves, especially higher-order ones, can represent a broad range of shapes, making them apt for capturing complex motion patterns.
\item Locality: Changes in a control point of a Bezier curve influence only a specific portion of the curve, ensuring that local modifications do not lead to global alterations in the trajectory shape.
\end{itemize}

\begin{figure}[!t]
\centering
\includegraphics[width=1\columnwidth]{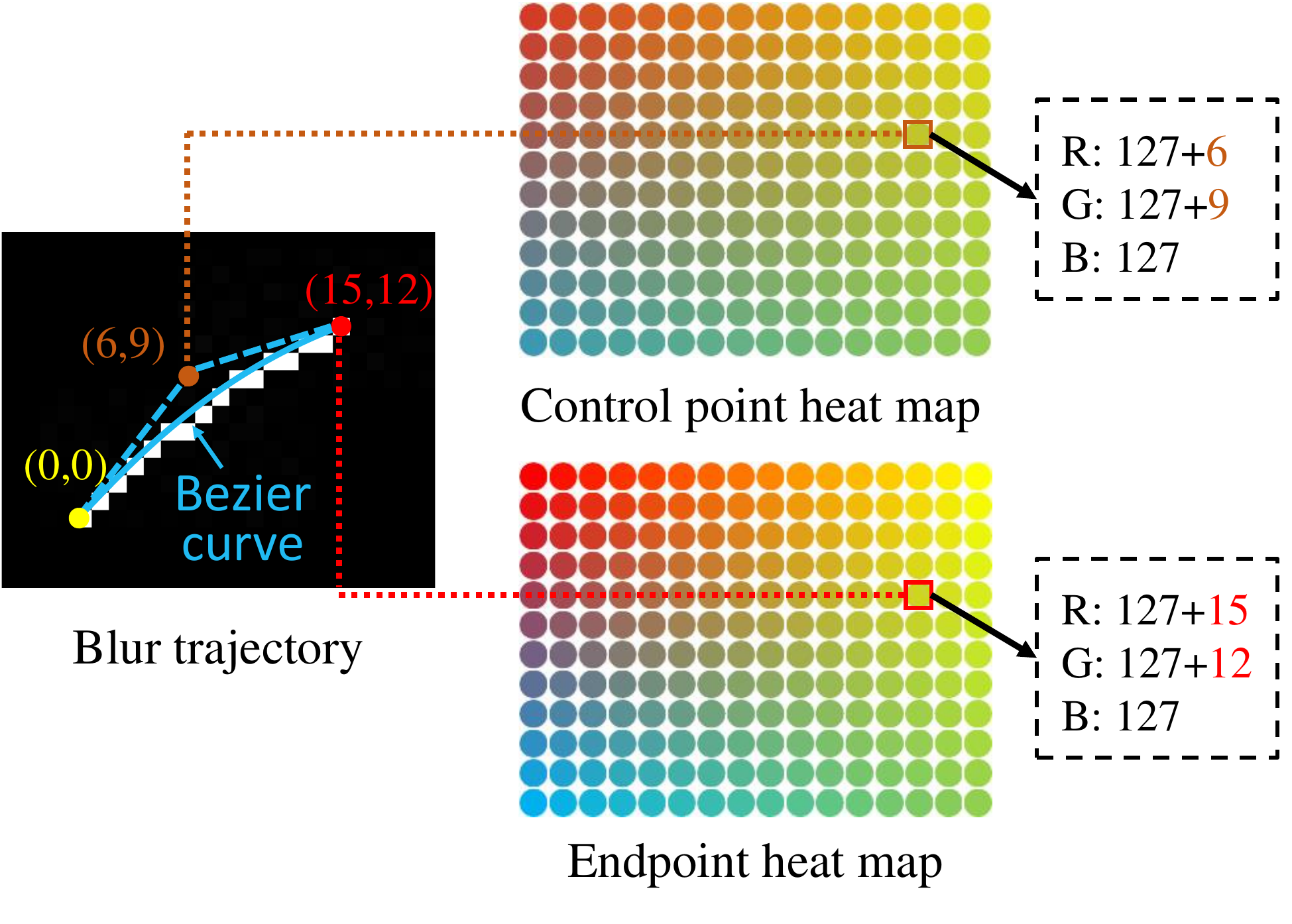} 
\vspace{-2em}
\caption{Estimation of a blur trajectory (white) with the Bezier curve (blue). The control points and endpoints are encoded with heat maps.}
\label{fig10}
\vspace{-1em}
\end{figure}

Notably, the start and endpoints of the blur trajectory (captured from gyroscope data) serve as the two main control points. Additional control points can be inferred from intermediate gyroscope readings, ensuring that curves closely follow blur paths. Our preliminary experiments show that Bezier curves' endpoints and control points can correctly approximate more than 99.9\% blur trajectories. The rest are sub-optimal but still tolerable in our proposed network. In addition, Bezier curves project different trajectories into precise length representations, providing more convenience for network training. During training and inference, together with blurred images, we project the endpoints and control points into heatmaps as auxiliary input features. The blur heat map is divided into a control point heat map and an endpoint heat map (see Fig.~\ref{fig10}).

\subsubsection{Network Architecture}
Fig.~\ref{fig3} illustrates the general idea of our proposed GAMD. It uses blurred images and heat maps as input. FPN (Feature Pyramid Network)~\cite{ghiasi2019fpn} is the main body for blurry image restoration. The rationale behind FPN is to restore low-level blur with the guidance of high-level context. While errors between the estimated and ground blur trajectories are sensitive (the higher the resolution, the smaller the error) to image resolution, we use the lower layer of the FPN for deblurring (with the guidance of the high-level features).

With FPN, GAMD is divided into three layers. The input feature groups of each layer are $C_1$, $C_2$, and $C_3$, and the output features are $P_1$, $P_2$, and $P_3$. $Rconv$ is the feature extraction module in the FPN~\cite{ghiasi2019fpn}, which is used for preliminary processing and fusion of each layer's blur images and heat maps. Note that the amount of information contained in the heat map is limited. To improve efficiency, we use $10\times 10$ convolutional layers to extract heat map features and $3\times 3$ convolutional layers to convolve the blurred image. Finally, these features are concatenated and fed into the $Rconv$ module. We use the $FFM$ module for feature fusion, which has been extensively used in segmentation and detection tasks~\cite{kirillov2019panoptic}. But different from traditional modules, we connect the output of the top layer to the $FFM$ of each layer, taking advantage of the top layer's context to improve the deblurring effect in the bottom layers. The $FFM$ module can be expressed as:
\begin{equation}
\begin{aligned}
    P_2 &= FFM(Rconv_2,Rconv_1)\\
    P_3 &= FFM((Rconv_3,Rconv_2),Rconv_1)
\end{aligned}
\quad .
\end{equation}
where $Rconv_n$ is the output of the n-th layer $Rconv$. This way, the network effectively reduces the noise from the bottom layer and fuses the features of adjacent layers. Finally, $Rconv$ is fed into the decoder to restore the details of the input image more accurately, thereby improving the restoration performance. Finally, we introduce the loss function. 
We use Euclidean loss between network outputs and the ground truth, and our Loss function is expressed as follows:
\begin{equation}
    Loss = \frac{1}{k_1}{\Vert}L_1-S_1{\Vert} + \frac{\omega}{k_3}{\Vert}L_3-S_3{\Vert}
\quad .
\end{equation}
where $L_n$ are GAMD output in the $n$-th scale, $S_n$ is the ground truth (downsampled to the same size using bilinear interpolation), $k_n$ is the number of elements to be normalized in $L_n$, $\omega$ is the weights for each scale, we set to 0.6.

\section{Experiments}
\label{s:exper}
In this section, we systematically evaluate the proposed blurred image synthesis method and validate the effectiveness and precision of GAMD on image deblurring tasks. Additionally, we discuss the limitations of our approach. We trained the network for 3000 epochs with a mini-batch size of 8 and an initial learning rate of 1e-4, decreased by a factor of 0.5 at every 500 epochs. Our experiments are performed on an NVIDIA RTX 3090, and the entire training process takes an average of 2 weeks.

\begin{figure*}[!t]
\centering
\includegraphics[width=1\textwidth]{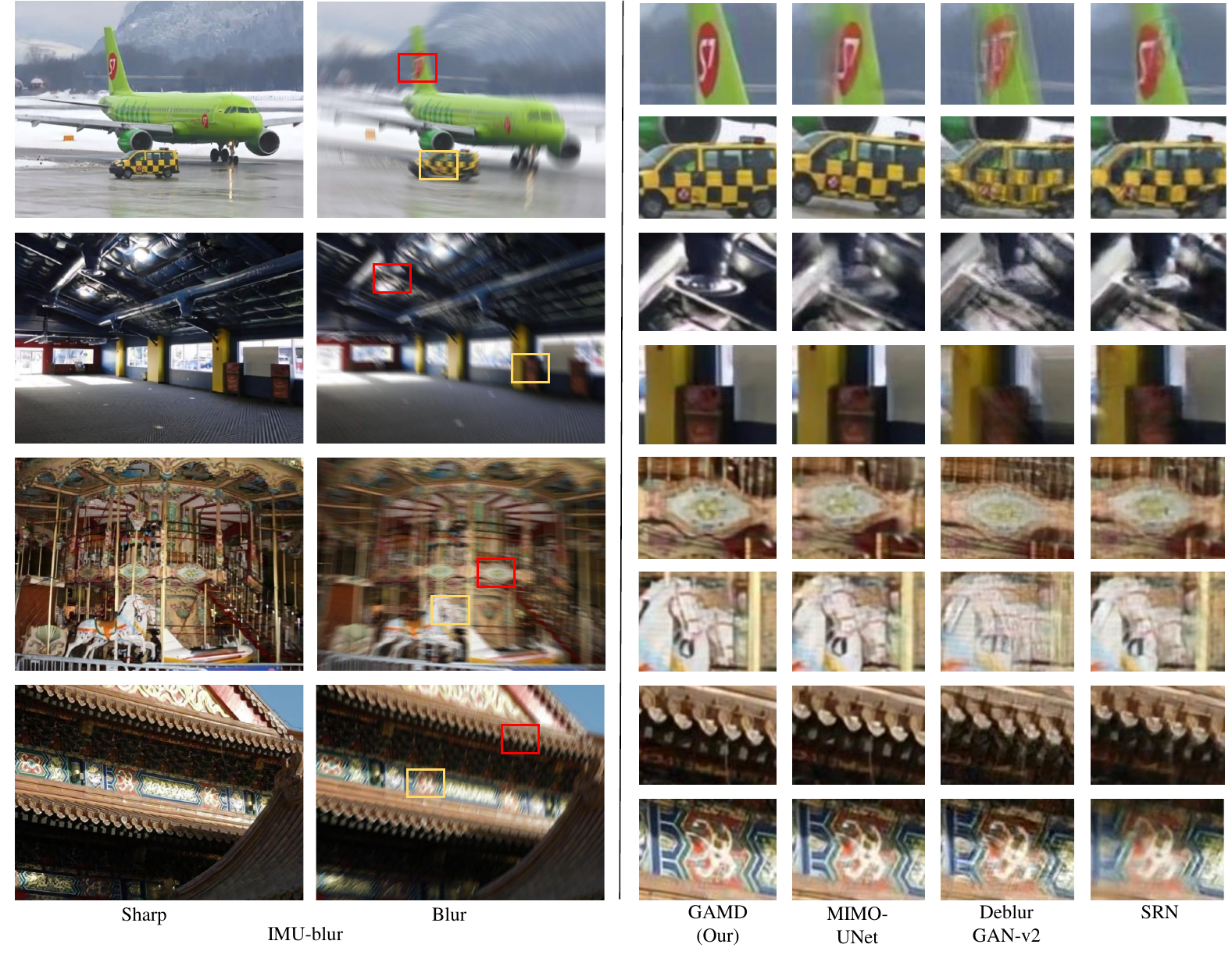} 
\vspace{-2em}
\caption{Illustration of the IMU-blur dataset (left) and the deblurring effects (right) from the corresponding marked regions: MIMO-UNet, DeblurGAN-v2, SRN and our proposed GAMD.}
\label{fig8}
\vspace{-1em}
\end{figure*}

\subsection{Datasets}
\noindent\textbf{IMU-blur}: We commenced our evaluation by randomly selecting 8350 clear images (aka. backgrounds) from existing image datasets~\cite{zhou2017places,quattoni2009recognizing}. By capturing IMU data during the motion blur induced by the D455i camera, we synthesized a dataset of 8350 blurred images accompanied by corresponding blur heat maps. Ultimately, this dataset, namely IMU-blur, contains 6680 triplets for training and 1670 triplets for testing. Our IMU-blur dataset has significant advantages over widely used datasets such as GoPro~\cite{nah2017deep} and RealBlur~\cite{rim2020real}, which contain limited scene variations. It includes images captured in different environments, eliminating the interference of features present in repeated scenes for network learning. Unlike existing blurred datasets mainly recorded by high-speed cameras, IMU-blur is more accessible to synthesize towards massive-produced and real-world motion blur. 

\noindent\textbf{BlurTrack}: We introduce BlurTrack, a dataset for evaluating synthetic blur methods to enhance dataset diversity. By projecting a matrix of laser dots onto the wall, we captured accurate blur trajectories using camera and IMU data. BlurTrack contains 3994 images, divided into different types to evaluate blurred synthetic scenes. BlurTrack allows us to assess the reliability of our synthetic blur methods and their agreement with real-world blur patterns.

\subsection{Blur Synthetic Evaluation}
To assess the effectiveness of our proposed synthetic strategy, we conduct two comparisons: (1) a comparison of blurred trajectories with those of natural images and (2) a comparison of trained deblurring networks with existing datasets. 

For the first one, we collect the starting point positions and corresponding IMU data of all blurred trajectories and use the method in this paper to predict the end point of the blurred trajectory. Comparing the expected blur track endpoints with those obtained from actual laser images (BlurTrack) allows us to quantify the difference in blurring effects. Our approach demonstrates an average pixel error of only two pixels compared to the exact blurred trajectory, indicating the high accuracy of the dataset. For the second one, we compare it with established datasets. We use this approach to synthesize new blurred datasets from clear images from the GoPro and RealBlur datasets. And trained models using the synthesized and original datasets. Subsequently, we evaluated their performance using SRN-DeblurNet~\cite{tao2018scale}. Table~\ref{tab2} presents a comparative performance analysis. Our synthetic dataset outperforms the original dataset, and notably, this is compared using synthetic and accurate data, which confirms the efficacy of our artificial blur dataset. Researchers can use this method to quickly generate large amounts of data sets, significantly reducing the difficulty of data set collection.

\begin{table}[!t]
\centering
\setlength{\tabcolsep}{5.2mm}{
\caption{Performance comparison of synthetic and current datasets. We train the network using our synthetic and original datasets and then simultaneously test the actual test set.}
\setlength{\tabcolsep}{5.5mm}{
\begin{tabular}{l|l|l}
    \hline
    Training sets & Test sets & PSNR/SSIM \\
    \hline
    GoPro & GoPro & 33.02/0.9265\\
    GoPro (Our) & GoPro & \textbf{33.14/0.9284}\\
    \hline
    RealBlur & RealBlur & 33.54/0.9093\\
    RealBlur (Our) & RealBlur & \textbf{34.00/0.9253}\\
    \hline
\end{tabular}}
\label{tab2}}
\vspace{-1em}
\end{table}

\subsection{Ablation Study}
We conduct experiments on IMU-blur and analyze the impact of blur heatmaps and deblurring networks on the deblurring effect, respectively. To conduct experiments more fairly, we use MIMO-UNet~\cite{cho2021rethinking} as the baseline method and improve it so that it can input blur heat maps. We call the improved MIMO-UNet~\cite{cho2021rethinking} as MIMO-Pro. First, we compared the effects of MIMO-Pro and MIMO-UNet~\cite{cho2021rethinking}. All parameters were set according to the original MIMO-UNet~\cite{cho2021rethinking}. The results are shown in the Table~\ref{tab3}. GAMD improves PSNR by 6.93 dB. We then compared the network without adding blur heat maps with MIMO-Unet~\cite{cho2021rethinking} and found that the PSNR increased by 1.73 dB. Experiments show that GAMD mainly improves the performance of the deblurring network by inputting additional blur information.

\begin{table}[!h]
\centering
\setlength{\tabcolsep}{5.2mm}{
\caption{In the ablation experiment, MIMO-UNet is the baseline method, and MIMO-Pro is based on the baseline method plus our blur heat map input. GAMD (Only Net) is based on GAMD by removing the blur heat map input.}
\setlength{\tabcolsep}{6.5mm}{
\begin{tabular}{l|l|l}
    \hline
    Model & PSNR & SSIM \\
    \hline
    MIMO-UNet & 24.39 & 0.7360\\
    MIMO-Pro & \textbf{31.32} & \textbf{0.9138}\\
    GAMD (Only Net) & 26.12 & 0.7830\\
    \hline
\end{tabular}}
\vspace{-1em}
\label{tab3}}
\end{table}

\begin{table}[!h]
\centering
\setlength{\tabcolsep}{7mm}{
\caption{Performance of different networks on IMU-blur.}
\vspace{-0.8em}
\begin{tabular}{l|l|l}
    \hline
    Model & PSNR & SSIM \\
    \hline
    SRN & 22.54 & 0.6604\\
    DeblurGAN-v2 & 21.45 & 0.6339\\
    MIMO-UNet & 24.39 & 0.7360\\
    \hline
    GAMD (Our) & \textbf{32.48} & \textbf{0.9014}\\
    \hline
\end{tabular}
\vspace{-1em}
\label{tab1}}
\end{table}

\subsection{Performance Comparison}
\label{s:exper:deblurnet}
We evaluated the GAMD against state-of-the-art deblurring networks, including SRN~\cite{tao2018scale}, DeblurGAN-v2~\cite{kupyn2018deblurgan}, and MIMO-UNet~\cite{cho2021rethinking}. GAMD exhibited superior performance, achieving a PSNR of 32.48 and an SSIM of 0.9014, significantly surpassing existing deblurring methods (Table~\ref{tab1}). Among traditional networks, MIMO-UNet~\cite{cho2021rethinking} showcased the best performance. Fig.~\ref{fig8} provides visual examples of the IMU-blur test dataset, comparing the deblurred images produced by GAMD (our method) against MIMO-UNet~\cite{cho2021rethinking}, DeblurGAN-v2~\cite{kupyn2018deblurgan}, and SRN~\cite{tao2018scale}. The presented results underscore the efficacy of our approach, particularly in mitigating the challenges of motion blur posed by significant camera shakes.

\subsection{Limitations}
\begin{figure}[!t]
\centering
\includegraphics[width=1\columnwidth]{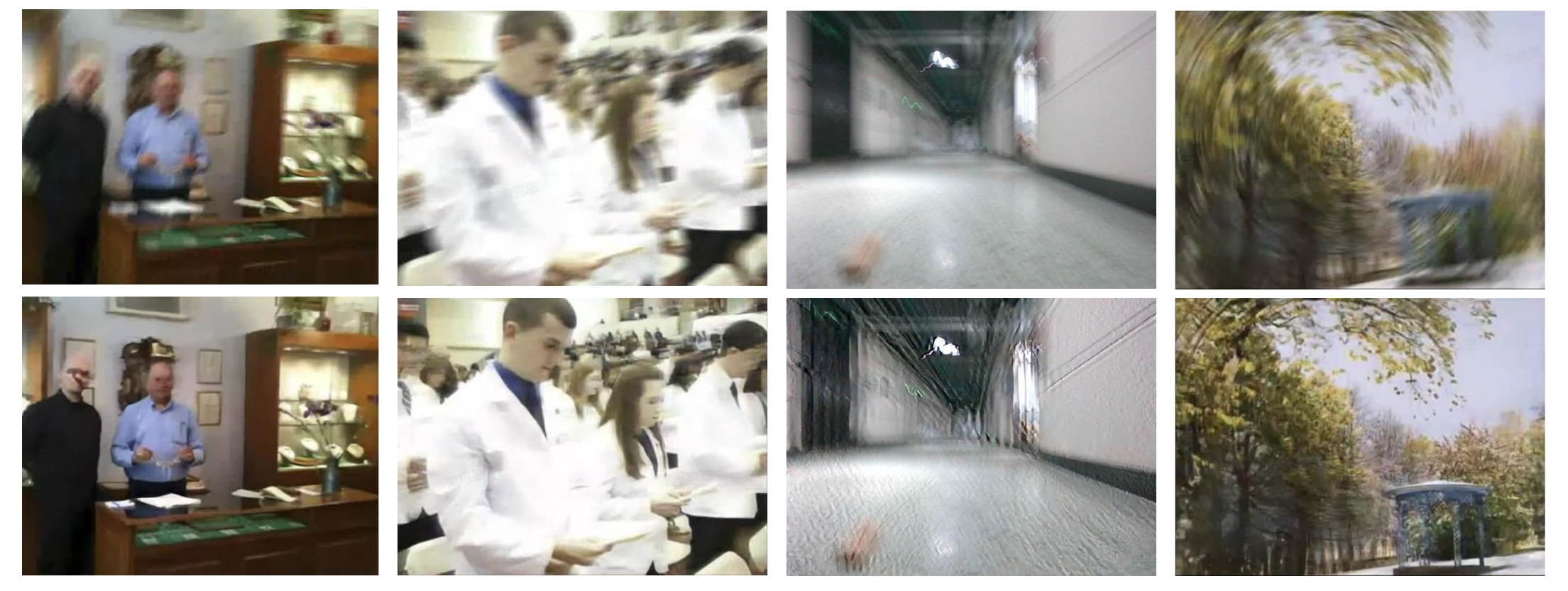}
\vspace{-1.5em}
\caption{Failure cases using our proposed GAMD. \textbf{Top}: Motion blur; \textbf{Bottom}: Restored with GAMD.}
\label{fig11}
\vspace{-1em}
\end{figure}

The proposed deblurring network may still struggle with significant camera roll and offset angles. For instance, Fig.~\ref{fig11} presents some failure cases with our GAMD. We can clearly observe the blurry effects, even after deblurring with our proposed GAMD. Nonetheless, GAMD and IMU-blur stand out for their superior deblurring performance, making them a contribution to the field.

\section{Conclusion}
\label{s:con}
This paper presents a novel approach for motion blur (caused by camera shake) synthesis and ultimately creates the blur dataset. At the same time, this paper also proposes a deblurring network, GAMD, which integrates blur trajectories, uses blur heat maps to collect blur trajectory information and guides image deblurring, enhancing network performance. Comparative experimental evaluation highlights the superiority of our approach over existing methods.

\bibliographystyle{named}
\bibliography{ming}

\end{document}